\documentclass{article}
\usepackage{ijcai21}

\usepackage{times}
\usepackage{soul}
\usepackage{url}
\usepackage[T1]{fontenc}    
\usepackage[hidelinks]{hyperref}
\usepackage[utf8]{inputenc}
\usepackage[small]{caption}
\usepackage{graphicx}
\usepackage{amsmath}
\usepackage{booktabs}
\usepackage{color}
\usepackage{amsthm}
\usepackage{multirow}

\usepackage{amssymb}
\usepackage{algorithm}
\usepackage{algorithmic}
\usepackage{tabularx}
\allowdisplaybreaks

\newtheorem{theorem}{Theorem}

\title{Bayesian Nonparametric Space Partitions: A Survey}
\author{
Xuhui Fan$^1$\and
Bin Li$^2$\and
Ling Luo$^3$\and
Scott A. Sisson$^{1}$\\
\affiliations
$^1$School of Mathematics and Statistics, University of New South Wales, Sydney\\
$^2$School of Computer Science, Fudan University\\
$^3$School of Computing and Information Systems, University of Melbourne\\
\emails
\{xuhui.fan, scott.sisson\}@unsw.edu.au,
libin@fudan.edu.cn, ling.luo@unimelb.edu.au
}

\begin{document}

\maketitle

\begin{abstract}
Bayesian nonparametric space partition (BNSP) models provide a variety of strategies for partitioning a $D$-dimensional space into a set of blocks, such that
the data within the same block share certain kinds of homogeneity. BNSP models are applicable to many areas, including regression/classification trees, random feature construction, and relational modelling. This survey provides the first comprehensive review of this subject.
We explore the current progress of BNSP research through  three perspectives: (1) {\em Partition strategies}, where we review the various techniques for generating partitions and discuss their theoretical foundation, `self-consistency'; (2) {\em Applications}, where we detail the current mainstream usages of BNSP models and identify some potential future applications; and (3) {\em Challenges}, where we discuss current unsolved problems and possible avenues for future research. 
\end{abstract}

\section{Introduction}
Describing and understanding the implicit and complex relationships between different features (covariates) in a dataset is typically a key modelling component of many advanced statistical and machine learning methods.
Among these approaches, Bayesian Nonparametric Space Partition (BNSP) models provide a flexible and geometrically interpretable way to describe these 
relationships. The main idea underlying BNSP models is to divide a $D$-dimensional ($D\ge2$) space into a number of `blocks', using a specified partition strategy, where 
the blocks are chosen such that the datapoints within in each block exhibit certain types of homogeneity. The resolution of the partition can be arbitrarily fine (i.e.~an arbitrary number of blocks in the space), perhaps as a function of the number of datapoints and the complexity of the relationship between the features, resulting in a flexible and adaptive nonparametric model able to describe complex data well.
Throughout, we define the random BNSP object, $\boxplus$, as a set of blocks,  $\boxplus=\{\Box_k\}^{k\in\mathbb{N}^+}$, where $\Box_k$ denotes the $k$-th block in the space. 
The space itself is commonly specified as $[0,1]^D$ with the observed data being rescaled to this domain.


As an illustration, consider the application of regression trees in credit risk assessment: the data $\{(\pmb{x}_n, y_n)\}_{n=1}^N$ are observed on $N$ individuals, where $\pmb{x}_n$ denotes individual $n$'s {attributes}~(e.g.~age, monthly expense, monthly income) and $y_n$ her {credit risk score}. In this setting, one block $\Box_k$ may be constructed as: [$20$, $25$] years old $\times$ [$\$2000$, $\$3000$] monthly expense $\times$ [$\$4000$, $\$4500$] monthly income. We have $\pmb{x}_n\in\Box_k$ if the $n$-th person belongs to the $k$-th block $\Box_k$. Individuals located in block $\Box_k$ are assumed to share similar behaviours in their risk scores. 

To construct regression trees based on BNSP models, an intensity variable $\omega_k$ is usually associated with each block $\Box_k$, such that $\Box_k$ contributes an impact with intensity $\omega_k$ to the labels $y_n$ of all  $\pmb{x}_n\in\Box_k$.
If we use a Gaussian distribution to describe the credit risk for individual $n$, the generative process could  be written as:
\begin{eqnarray}
   (1)\quad & \{\Box_k\}_k & \sim  \text{BNSP}([0, 1]^D, -)\nonumber \\
   (2)\quad & \{\omega_k\}_k & \sim  \mathcal{N}(0, \delta^2) \nonumber \\
   (3)\quad & y_n & \sim  \mathcal{N}(\sum_k\omega_k\cdot \pmb{1}_{\pmb{x}\in \Box_k}(\pmb{x}_n), \sigma^2)\nonumber
\end{eqnarray}
where $\pmb{1}_{\pmb{x}\in \Box_k}(\pmb{x}_n)=1$ if $\pmb{x}_n\in\Box_k$; otherwise $\pmb{1}_{\pmb{x}\in \Box_k}(\pmb{x}_n)=0$. Step $(1)$  generates the blocks from a BNSP model (with additional parameters denoted by `$-$') on the space spanned by the (rescaled) feature data $\{\pmb{x}_n\}_{n=1}^N$. This step is the primary research focus for BNSP models -- {\em the investigation of efficient ways to generate meaningful blocks in the space}. Step $(2)$ generates the intensity values $\{\omega_k\}_k$ for all blocks from a common Gaussian distribution. Step $(3)$ generates the label data from a Gaussian distribution, with mean given by the sum of intensities of all blocks covering $\pmb{x}_n$, and with error variance $\sigma^2$. Through posterior inference on these random variables (e.g., the blocks $\{\Box_k\}_k$, intensities $\{\omega_k\}_k$, and error variance $\sigma^2$), which is typically implemented using Markov chain Monte Carlo (MCMC) methods, insightful structural information can be uncovered.

With different settings of the likelihood function for the label data, regression trees can be built for either regression or classification tasks. E.g.,  the Bernoulli distribution can be used as the likelihood function, with probability given by the logistic transform of the sum of intensities. Similarly a categorical distribution can be used, with the probability vector given by a multinomial logistic transform of the sum of intensities.

In contrast to the Dirichlet process~\cite{ferguson1973bayesian} and its explicit view, the stick-breaking process~\cite{sethuraman1994constructive}, which are often applied only in one-dimensional-space problem settings, 
BNSP models are typically implemented in multi-dimensional spaces. 
For particular definitions of the space and modelling objectives, BNSP models have been successfully implemented in many real-world applications. For example, when the space is spanned by data features and the modelling objective is the labels attached to those features, BNSP models can be applied to regression/classification trees~\cite{chipman2010bart,LakRoyTeh2014a} and online learning~\cite{LakRoyTeh2014a,lakshminarayanan2016mondrian,consistencyMondrianforest,fan2020onlinebspf}; when the space is spanned by the communities in a social network  and the modelling objective is the linkages between the nodes in the network, BNSP models can be implemented in relational modelling~\cite{kemp2006learning,roy2009mondrian,nakano2014rectangular} and community detection~\cite{nowicki2001estimation}. Other examples of applications include random feature construction~\cite{balog2016mondrian}, voice recognition~\cite{mhmm2014nakano}, co-clustering~\cite{wang2011nonparametric}, and matrix permutation approximation~\cite{NIPS2019_9089}.
Some potential application areas of BNSP models are spatial-temporal modelling, and image detection and segmentation.

In this survey, we will explore the following aspects of BNSP models: (1) {\em Partition strategies}, where we first introduce an important property in defining the model -- self-consistency -- and then review the various strategies that have been developed for generating the partitions in the space. These include grid-style partitions, hierarchical partitions~(axis-aligned cuts), hierarchical partitions~(sloped cuts), floorplan partitions, and rectangular bounding partitions. We also summarize and contrast the characteristics of each approach as no single strategy will dominate the others in all cases; (2) {\em Applications},  where we demonstrate how to apply BNSP models to the real-world studies of online learning, random feature construction and relational modelling; (3) {\it Challenges}, where we discuss the current challenges in BNSP research and avenues for future work, including scalable inference methods, partition flexibility, posterior concentration analysis and analysis of deep neural networks.

\section{Partition strategies}
\begin{figure*}[t]
\centering
\includegraphics[width = 0.75 \textwidth]{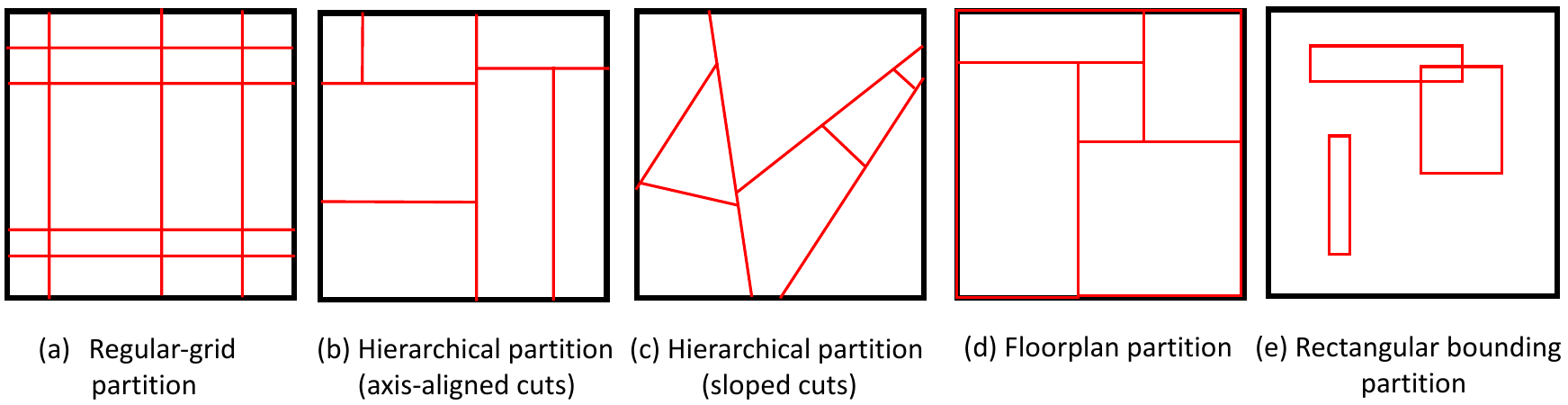}
\caption{Visualisation of five typical BNSP partition strategies in a $2$-dimensional space.}
\label{cutting_vis}
\end{figure*}

The partition strategies for BNSP models can be divided into 5 categories. Illustrations of particular implementations of these in 2-dimensional space are shown in Figure~\ref{cutting_vis}. A critical aspect underpinning BNSP processes is the concept of {\em self-consistency} of the process. We review this below.

\begin{figure}[t]
\centering
\includegraphics[width = 0.45 \textwidth]{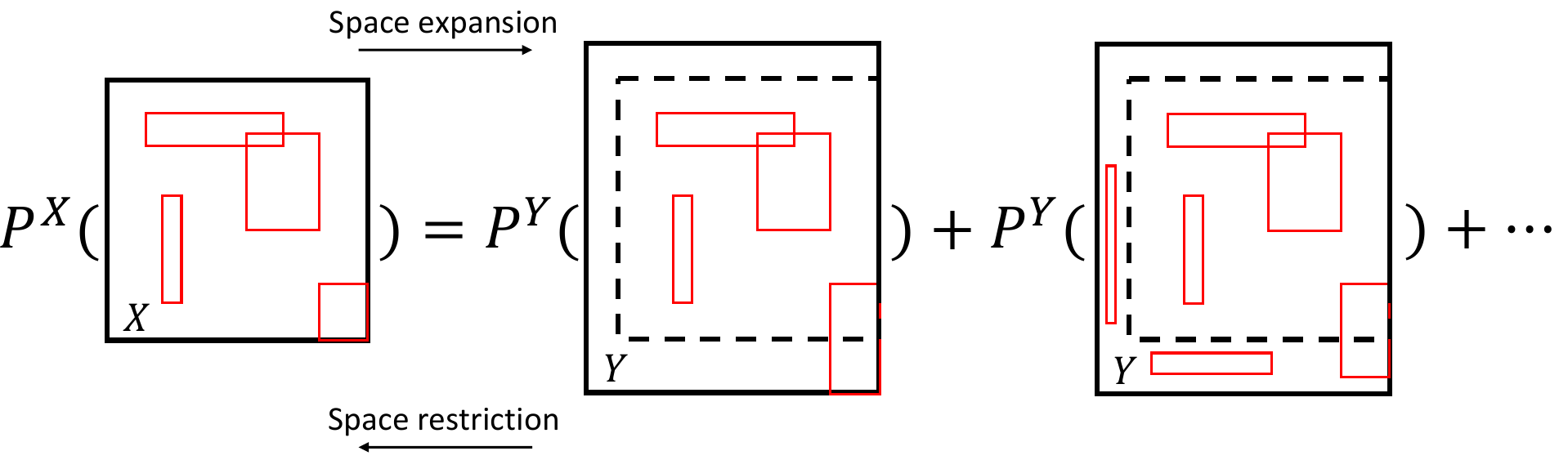}
\caption{Illustration of self-consistency of the Rectangular Bounding Process (RBP). The solid black boxes denote $2$-dimensional spaces; the red boxes denote different blocks generated by the RBP. The {\it lhs} of the equality denotes the probability of an RBP partition in a smaller space $X\subset Y$, and the {\it rhs} shows the probabilities of an infinite number of possible partitions expanded to a larger space $Y$ from $X$ (i.e.~the {\it lhs} probability is obtained by marginalizing out all possible partitions in the expanded region $Y/X$).}
\label{self_consistency:rbp}
\end{figure}

\vspace{0.2cm}
\noindent \textbf{Self-consistency:}
The main idea behind self-consistency is to ensure that the probability of any partition remains invariant under the expansion or shrinkage of the space.

More formally, suppose we have a partition $\boxplus_Y$ of a space $Y$, generated from a BNSP process, where $Y \in \mathcal{F}(\mathbb{R}^D)$  denotes the set of all finite rectangular boxes in $\mathbb{R}^D$.
When restricting the BNSP process to a sub-space $X$, where $X \subset Y$, the resulting partition of $\boxplus_Y$ restricted to $X$ should be distributed as if it were generated via the BNSP process working on $X$ directly. Figure~\ref{self_consistency:rbp} illustrates the self-consistency of the Rectangular Bounding Process (RBP)~\cite{NIPS2018_RBP}, which will be reviewed in detail below (Section \ref{sec:RBP}). A BNSP process is defined to be self-consistent if and only if the probability of a partition on $X$ equals the sum of probabilities of all possible partitions on $Y$ extended from the partition on $X$.
Mathematically, this can be represented as $P^Y_\boxplus\left(\pi_{Y,X}^{-1}(\boxplus_X)\right) = P^X_\boxplus(\boxplus_X)$, where $\pi_{Y,X}$ denotes the restriction operator from $Y$ to $X$. For continuous spaces, the number of possible extensions is usually uncountably infinite.

The self-consistency property ensures that the partition of a smaller space can be safely extended to larger spaces. Online learning is a typical scenario in which self-consistency is essential. When new data points are observed outside the range of the current data (i.e~the minimum bounding box of all existing data points, $X$), the BNSP sample can immediately be extended to include these new data points within a new bounding box ($Y$). Further, if we make use of the Kolmogorov Extension theorem~\cite{oksendal2013stochastic}, the self-consistency property then allows BNSP models to define partitions in {\em infinite} multi-dimensional space.

We now explore the various partition strategies, where the order is approximately in terms of proposed time and model complexity of these strategies.

\subsection{Regular-grid partitions}
Grid-style BNSP models use one-dimensional partitions on each univariate dimension to construct a regular-grid on the full dimensional space.
%
Regular-grid partitions 
extend trivially from one-dimensional partitions (e.g.~the stick-breaking process), and are a historically earlier and simpler instance of BNSP models. For a $D$-dimensional space, the model is constructed using $D$ independent stick-breaking processes, each designated as a partition of one dimension. The orthogonal crossover of the partition on these dimensions (Figure~\ref{cutting_vis}(a)) produces regular-grid blocks in the full-dimensional space. Due to the self-consistency of the stick-breaking process in each dimension separately, the resulting regular-grid partition is also self-consistent.

The regular-grid partition is an over-simplified partition strategy, as its construction ignores any dependency between dimensions and generates partitions over each dimension independently. Accordingly, this approach is very likely to generate trivial and unneeded blocks in data-sparse regions. Currently, this approach is mainly  applied in relational modelling in $2$-dimensional space. A specific implementation is the Infinite Relational Model (IRM)~\cite{kemp2006learning}, which is an infinite-state variant of the Stochastic Block Model (SBM)~\cite{nowicki2001estimation}. Other models in this category include dynamic IRMs~\cite{ishiguro2010dynamic} for modelling dynamic relational data, and multi-membership relational modelling~\cite{nonpa2013schmidt}.

\subsection{Hierarchical partitions~(axis-aligned cuts)}
Hierarchical partitions follow a top-down strategy to recursively cut an existing block into two new blocks. In this way, the blocks are organized in a binary tree. The Mondrian process (MP)~\cite{roy2009mondrian} is the representative construction of hierarchical partitions. In general, the MP recursively generates axis-aligned cuts on a unit hypercube $[0,1]^D$, and divides the space in a hierarchical fashion known as $k$d-tree (Figure~\ref{cutting_vis}(b)). The $k$d-tree construction process is regulated by attaching an exponentially distributed cost to each axis-aligned cut. The process is then terminated when the accumulated sum of the costs exceeds a provided budget value. Because the MP can partially consider inter-dimensional dependency, it can produce fewer trivial blocks than regular-grid partitions. The MP retains the self-consistency property by carefully modelling the relationship between the cut cost and the generation of axis-aligned cuts.

\subsection{Hierarchical partitions~(sloped cuts)}
In comparison to the axis-aligned cuts, a related group of tree-structured hierarchical partition models consider sloped (non-axis-aligned) cuts for cutting the space (Figure~\ref{cutting_vis}(c)). BNSP processes adopting this strategy include the Binary Space Partition-Tree (BSP-Tree) process~\cite{pmlr-v84-fan18b}, and Random Tessellation Forests (RTF)~\cite{random_tessellation_forests}. In contrast to the axis-aligned partitions of the MP, the BSP-Tree process accounts for inter-dimensional dependency by generating sloped cuts, and thereby forming convex polygon-shaped blocks in a $2$-dimensional space. Binary Space Partition Forests~\cite{pmlr-v89-fan18a} extend the mechanism of producing sloped cuts from $2$-dimensional spaces to $D$-dimensional spaces ($D\ge2$). This results in the production of convex-polyhedron blocks, with the restriction that the cutting hyperplane is parallel to $D-2$ dimensions. 

Similarly, Random Tessellation Forests  are constructed by generating arbitrary sloped cutting hyperplanes in $D$-dimensional space. These sloped cuts permit a greater focus on describing the multi-dimensional dependence, and as a result, the model has the potential to produce partitions more efficiently in the space. Recent work is able to efficiently generalise cut directions  using the iteration stable (STIT) tessellations technique~\cite{o2020stochastic}. 
\subsection{Floorplan partitions}
It is noted that the regular-grid partitions and hierarchical partitions~(axis-aligned cuts) are limited to support a subset of floorplan partitions~(e.g., they cannot form Figure~\ref{cutting_vis}(d)). Baxter Permutation Partitions~(BPPs)~\cite{NEURIPS2020_6271faad} were recently proposed to produce arbitrary floorplan partitions on a continuous space (Figure~\ref{cutting_vis}(d)). The BPP first uses the one-to-one correspondence between Baxter permutations~\cite{DULUCQ1998143} and floorplan partitions to generate floorplan partitions on $2D$ arrays. This is then combined with the block-breaking process -- a multi-dimensional extension of the stick-breaking process~\cite{sethuraman1994constructive} -- to produce a discrete-time Markov process over the set of floorplan partitions on a continuous space. By getting rid of the restriction of a regular-grid or hierarchical structure, the BPPs are able to produce a variety of partition structures (Figure~\ref{cutting_vis}(d)). The BPPs are self-consistent and have been successfully used in the relational modelling setting. 

A discrete space alternative is the Rectangular Tiling Process~(RTP) \cite{nakano2014rectangular}, which produces a floorplan partition structure on a $2$-dimensional array. It typically uses the geometric distribution to generate the length of each block, with the constraint that the length does not violate the rectangular restriction of  existing blocks. It is also self-consistent due to the memoryless property of the geometric distribution. However, its generative process is quite complicated for practical usage and the domain of discrete array in RTP restricts it being used in e.g.~relational modelling applications only. In contrast, the partition strategies discussed above can be applied to both continuous space and multi-dimensional arrays (with trivial modifications).

\subsection{Rectangular bounding partitions}
\label{sec:RBP}
In direct contrast to the cutting-based strategies (including grid-style partitions, hierarchical partitions and floorplan partitions), the Rectangular Bounding Process (RBP)~\cite{NIPS2018_RBP} uses a bounding strategy to partition the space. By independently constructing rectangular bounding boxes in the space, the RBP can cover and concentrate on significant regions and avoid data sparse regions (Figure~\ref{cutting_vis}(e)). The RBP requires a budget parameter $\tau$ to control the number of bounding boxes,
and a length parameter $\lambda$
to control the size and location of the bounding boxes. The probability that any data point occupies a box is the same. The RBP is  self-consistent, and can generate more bounding boxes given a larger budget or a larger space. 

The biggest advantage of the RBP is the parsimonious partition of space. Real data is typically not evenly distributed over the entire space, but clusters in local regions. The cutting-based strategies, such as grid-style and hierarchical partitions, inevitably produce too many cuts for sparse regions with few data points while trying to fit data in the dense regions. Recall the credit risk modelling problem, where two dimensions of the feature space are `age' and `salary'. Traditional cutting-based models may inevitably cut the regions of young age and very high salary even if there are very few people in those regions, simply by placing cuts in other areas of the space. In contrast, the RBP can place more bounding boxes in the most important regions, and fewer in data-sparse and noisy regions.
As a result, the RBP is able to balance the \emph{fitness} and \emph{parsimony} of the partition.

\begin{table*}[t]
\caption{\small Comparison of various space-partition strategies (IDD: Inter-Dimensional Dependency; RM: Relational Modelling; RT: Regression Trees; RFC: Random Feature Construction; PG: Particle Gibbs; ICM: Iterative Conditional Modes).}
\label{discussion_on_partitions}
\centering
{\small\begin{tabular}{|c|cccccc|}
  \hline
{Models} & Self-consistency & Continuous & $D$ & Inference & IDD & Applications
\\  
\hline
Regular-grid & $\checkmark$& $\checkmark$ & $2$ & Gibbs &  $\times$ & RM\\
MP & $\checkmark$ & $\checkmark$ & $\ge 2$ & RJ-MCMC \& PG & $\checkmark$ & RM \& RT \& RFC  \\ 
BSP & $\checkmark$ & $\checkmark$ & $\ge 2$ & RJ-MCMC \& PG & $\checkmark$ & RM \& RT  \\ 
RTF & $\checkmark$& $\checkmark$ & $\ge 2$ & SMC & $\checkmark$ & RT\\
BPP & $\checkmark$& $\checkmark$  & $ 2$ & M-H & $\checkmark$ & RM\\
RBP & $\checkmark$& $\checkmark$ & $\ge 2$ & M-H & $\times$ & RM \& RT\\
\hline
Plaid & $\times$  & $\times$ & $2$ & Gibbs & $\times$ & RM\\
MTA & $\times$ & $\times$ & $2$ & ICM & $\times$ & RM\\
BART & $\times$ & $\checkmark$ & $\ge 2$ & Gibbs & $\times$ & RT\\
\hline
\end{tabular}}
\end{table*}

\subsection{Non-self-consistent partitions}
There are a number of space partition models that do not have the self-consistency property, which implies that their applicability is limited. 
Bayesian plaid models~\cite{caldas2008bayesian} generate `plaid'-like (discrete) partitions on a $2$-dimensional array. Usually, the plaids are generated through the Beta-Bernoulli process, or the Indian Buffet Process (IBP) for an infinite number of plaids. 
The plaid models can produce bounding boxes similar to the continuous-space RBP, although the plaid models are restricted to discrete spaces. Each block is formed through individual row and column permutations. Plaid models are not self-consistent.

Similar to Bayesian plaid models, Matrix Tile Analysis (MTA) models~\cite{Givoni06} are constructed on a $2$-dimensional array, although it is a non-Bayesian method. MTA generates rectangular boxes on discrete arrays with the constraint that the boxes cannot be overlapped. 
Due to its non-Bayesian nature, the self-consistency property is not applicable to it.

Bayesian Additive Regression Trees (BART)~\cite{chipman2010bart} is a space-partition model using the hierarchical partition strategy, but without the self-consistency property. In general, BART assigns uniform distributions to the cutting positions and uses the Bernoulli distribution to regulate the tree depth. As the parameter in the Bernoulli distribution is inversely related to the depth of the node in the tree, deeper nodes have a lower probability of being split.

\subsection{Comparison}
Table~\ref{discussion_on_partitions} summarises the similarities and differences of the reviewed space-partition processes, in terms of: whether it possesses self-consistency, whether it is applicable to continuous or discrete spaces, the number of dimensions $D$ in which it can be applied, the available inference methods, whether it can account for inter-dimensional dependency, and suitable applications. As is apparent, there is not one single model that can dominate the others in all modelling aspects, so in practice it is important to select the most appropriate technique.

Most of the inference algorithms for BNSP models are based on Markov chain Monte Carlo (MCMC) methods. In particular, Gibbs samplers are used for the grid-style partition models, Bayesian plaid models and BART, as conjugacy  between the prior and posterior distributions is satisfied by each latent variable. The Reversible-Jump MCMC (RJ-MCMC) and Particle Gibbs (PG) algorithms are required for the hierarchical partition-based  models, due to the tree structures of their latent variables. The Metropolis-Hastings (M-H) algorithm is used for the RTP, BPP and RBP models, due to the specially designed distributions of their latent variables. The Iterative Condition Modes (ICM) method~\cite{Givoni06} is used for the non-Bayesian MTA model.

\section{Applications}
In addition to  regression trees (with static data), BNSP constructions can be applied to other modelling scenarios.
Here we review three of these: online learning, random feature construction, and relational modelling.

\begin{figure*}[ht]
\centering
\includegraphics[width = 0.65\textwidth]{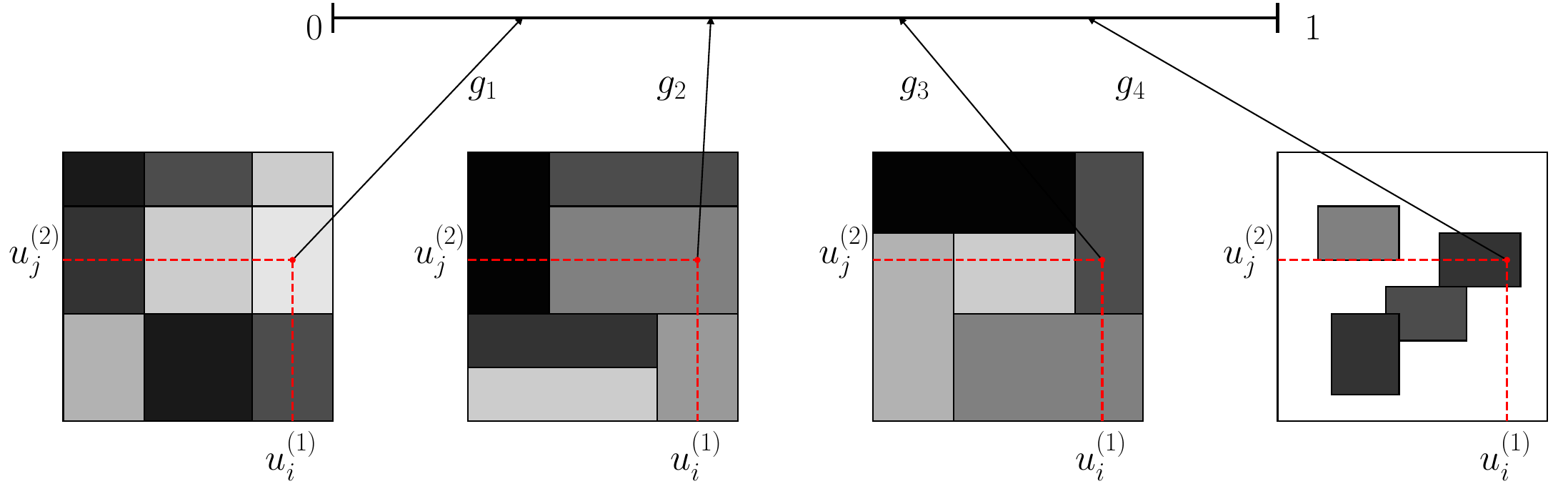}
\caption{Relational modelling using BNSP models. The relation intensity of any pair of coordinates can be mapped to a unit interval using a graphon function (denoted by $g_1,\ldots,g_4$). From left to right: regular-grid partitions; hierarchical partition~(axis-aligned cuts); floorplan partitions, and rectangular bounding partitions. The darker color of the point $(u_i^{(1)},u_j^{(2)})$ corresponds to a higher intensity in the interval, which means that there is a higher probability to generate a link between $u_i^{(1)}$ and $ u_j^{(2)} $. }
\label{fig:graphon_comparison}
\end{figure*}

\subsection{Online learning} 
Some BNSP processes can be extended beyond regression-tree modelling for static datasets, to the online learning setting~\cite{LakRoyTeh2014a,lakshminarayanan2016mondrian,consistencyMondrianforest,fan2020onlinebspf}. 
Suppose we observe a set of $N$ labeled data points $\{(\pmb{x}_n, y_n)\}_{n=1}^N\in\mathbb{R}^D\times \mathbb{R}$ which arrive over time, with $y_n$ as the corresponding label of $\pmb{x}_n$. When a new data point arrives, it is incorporated into the analysis, and the BNSP structure is updated accordingly without refitting the model using the full dataset.

The online learning application of BNSP processes follows the same spirit as online random forest-type algorithms~\cite{breiman2000some}, which assume that the tree-structured model is generated independently of the data labels. The Mondrian Forest (MF)~\cite{LakRoyTeh2014a,lakshminarayanan2016mondrian} is the first model to apply BNSP processes to the online learning setting. It uses the Mondrian process~\cite{roy2009mondrian} to place a probability distribution over all $k$d-tree partitions of the space. To regularize the MF to be universally consistent~(which ensures the prediction error converges to the Bayes error), the budget parameter is increased with the amount of data~\cite{consistencyMondrianforest}, and so the model can achieve the minimax rate in a multi-dimensional space for single decision trees. \cite{mourtada2018minimax} shows the advantage of forest-based settings through improved convergence results. The Online Binary Space Partition Forest~\cite{fan2020onlinebspf} similarly extends the BSP-Tree to the online learning setting, randomly generating sloped hyperplanes to cut the feature space.

\subsection{Random feature construction}

The generative process of BNSP models can support the construction of random features to approximate kernels~\cite{rahimi2008random}. Given $N$ data points and $K$ generated blocks, a binary random feature matrix $\pmb{\Phi}\in\{0, 1\}^{N\times K}$ can be used to record the box coverage status for the data points, where the $(n,k)$-th entry indicates whether the $n$-th data point is covered by the $k$-th block. The random feature $\pmb{\Phi}_n$ for each data point can replace the observed feature vector $\pmb{x}_n$, and be used to approximate certain kinds of kernel. 

The Mondrian Kernel (MK)~\cite{balog2016mondrian} was the first model to implement the random feature construction based on BNSP models. In particular, the MK generates a set of $M$ Mondrian process partitions on the feature space and incorporates all $M$ partitions into one random feature matrix $\bar{\pmb{\Phi}}\in\{0, 1\}^{N\times (MK)}$. As a result, the product {$\bar{\pmb{\Phi}}_{n}\bar{\pmb{\Phi}}_{n'}^{\top}$} represents the count of blocks covering both the $n$-th and $n'$-th data points. \cite{balog2016mondrian} shows that both the expectation  of {$\bar{\pmb{\Phi}}\bar{\pmb{\Phi}}^{\top}$}, and the case of {$\bar{\pmb{\Phi}}\bar{\pmb{\Phi}}^{\top}$} as $M\to\infty$ results in the Laplace kernel.  \cite{o2020stochastic} extends the axis-aligned cuts in the Mondrian Process to arbitrary directions, and characterise all possible kernels~(including the radial basis function kernel and the Laplace kernel) that the hierarchical partitions can approximate.

\subsection{Relational modelling for link prediction}
Each data point in relational modelling represents the linkage information of entities, with the value of the link (e.g.~$R_{ij}$ between individuals $i$ and $j$) as the label. Each entity is associated with one latent covariate (e.g.~an {\it a priori} uniformly distributed variable $u_i$ for individual $i$). The feature information for $R_{ij}$ is then the pair of random variables $(u_i, u_j)$.
The blocks then define community-by-community interactions, and the links located in the same block  follow the same Bernoulli distribution.

For relational modelling, the observed linkage data $\pmb{R}=[R_{ij}]$ is regarded as the label data, which is usually presented as a symmetric (undirected) or asymmetric (directed) matrix $\pmb{R} \in \{0,1\}^{N \times N}$, with $R_{ij}=1$ indicating that entity $i$ interacts with entity $j$, otherwise $R_{ij}=0$. Since $\pmb{R}$ represents pairwise linkage relations, the space of this application is spanned by two community distributions, and encoded as a unit square $[0, 1]^2$. For the feature data of each linkage $R_{ij}$, the model generates a {\it pseudo} attribute $u_i^{(1)},u_i^{(2)}$ for each entity $i$ at {\it pseudo} features $1,2$, and concatenates attributes~(i.e. $[u_i^{(1)}, u_j^{(1)}]^{\top}$) for entities $i$ and $j$ as these {\it pseudo} features.

We can summarize the generative process of relational modelling based on a BNSP model as: 
\begin{eqnarray}    
(1)\quad &\{\Box_k\}_k & \sim \text{BNSP}([0, 1]^2, -);\nonumber \\
(2)\quad&\{\omega_k\}_k & \sim \text{Beta}(-);
\nonumber \\
(3)\quad&\{u_i^{(1)},u_i^{(2)}\}_i & \sim\text{Uniform}[0,1];\nonumber \\
(4)\quad &R_{ij} & \sim \text{Bernoulli}(\omega_{k:{(u_i^{(1)}, u_j^{(2)})\in \Box_k}}).\nonumber
\end{eqnarray}
where the set of blocks $\{\Box_k\}_k$ represents meaningful community-by-community interaction groups, and the intensity $\omega_k$ denotes the influence contributing to the links belonging to $\Box_k$. 

There are various extensions to the above basic generative process. In the case of overlapping blocks~\cite{NIPS2018_RBP}, the intensity can be assumed to follow a Normal distribution. Positive (negative) intensity would then promote (suppress) links located in the corresponding block. In the case of categorical or real-valued links, different types of likelihood functions can be adopted. E.g., a Gamma distribution likelihood can be used to model non-negative real-valued links.

\vspace{0.2cm}
\noindent \textbf{Connection to graphons:} Relational modelling applications are closely related to the graphon (graph function) literature~\cite{TPAMI2014peterdaniel}. Given the exchangeable relational data for relational modelling, the Aldous--Hoover theorem~\cite{aldous1981representations} provides the theoretical foundation to model exchangeable multi-dimensional arrays conditioned on a stochastic partition model. A random $2$-dimensional array is separately exchangeable if its distribution is invariant under separate permutations of rows and columns. Specifically, the theorem states that:
\begin{theorem}~\cite{TPAMI2014peterdaniel} \label{theoremgraphon}
  A random array $(R_{ij})$ is separately exchangeable if and only if it can be represented as follows: there exists a random measurable function $F:[0,1]^3 \mapsto \mathcal{X}$ such that $(R_{ij}) \overset{d}{=} \left(F(u_i^{(1)}, u_j^{(2)}, \nu_{ij})\right)$, where $\{u_i^{(1)}\}_i, \{u_j^{(2)}\}_j$ and $\{\nu_{ij}\}_{i,j}$ are two sequences and an array of i.i.d.~uniform random variables in $[0,1]$, respectively.
\end{theorem}

Many of the BNSP models comply with this theorem, with specific forms of mapping function $F$. For instance, as illustrated in Figure~\ref{fig:graphon_comparison}, given the uniformly distributed node coordinates $(u_i^{(1)}, u_j^{(2)})$, the regular-grid partition is related to a regular-grid graphon; the hierarchical partitions~(axis-aligned cuts) and hierarchical partitions~(sloped cuts) are related to $k$d-tree structured graphons; the floorplan partition is related to a guillotine partition graphon; and the RBP is related to a box-wise constant graphon. All these graphons are piece-wise constant in $[0,1]^2$.

%

\section{Challenges}


While BNSP models have developed into highly flexible models that have demonstrated outstanding success in a number of useful applications, there are still a number of open research questions and remaining challenges.

\subsection{Scalable inference methods}
To the best of our knowledge, most  inference methods for BNSP models rely on MCMC simulation (see  Table~\ref{discussion_on_partitions}). Since MCMC methods often result in long computational times and require convergence assessment, alternative scalable inference methods are necessary to deal with larger-scale data problems. Variational methods, in particular the popular variational auto-encoder methods~\cite{kingma2013auto}, are promising solutions as they produce optimization-based (rather than simulation-based) posterior approximations. However, there has been little progress in developing variational inference methods for BNSP models, with the exception of the regular-grid partition. One possible reason for this might be the complications of the partition structure, such as the tree structure in a hierarchical partition.

Within hierarchical partition models, the BNSP strategy of using sloped cuts for partitions seems to be more effective and flexible than the axis-aligned cutting strategy. However, the improvement of modelling capability comes at the price of increased computational cost. The cost of current sloped-cut models scales at least quadratically with the number of dimensions of the space, whereas the cost of axis-aligned models often scales linearly with the number of dimensions. Efficient ways of circumventing computational complexity, while retaining self-consistency, is an interesting challenge.

\subsection{Flexibility of Partitions}
\noindent \textbf{Dependent BNSP partitions:} Currently, BNSP models generate space partitions independently of other partitions. One potential for improvement could be to allow for dependence between different partitions, and accordingly extend the application of BNSP models to a greater range of data formats, such as dynamic data, in which data has dependence across different time points, cross-domain data, in which data may contain dependencies across different domains, or multi-view data, in which data contains dependencies through different descriptions. However, such extensions would be nontrivial, as it would be easy to violate self-consistency. Innovative approaches may be needed to incorporate partition dependence while simultaneously retaining self-consistency. 

\vspace{0.2cm}
\noindent \textbf{Convex-polygon bounding blocks:} Within bounding block-based strategies, it is assumed  that the box construction is independent in all dimensions, and so rectangular blocks are produced. In practice, the most efficient shape of a block might be convex or even an irregular polygon. For example, in credit risk modelling (with expense and salary), one high risk block might be formed by a sloped cutting line of expense $\ge$ salary. In this problem, the challenge is to define the convex polygon while retaining self-consistency.

\subsection{Posterior concentration analysis}
The importance of posterior concentration behavior in the Bayesian generative process has been repeatedly emphasized in the literature. Currently, \cite{rockova2017posterior} has developed a series of posterior concentration results for BART, based on the work of~\cite{ghosal2007convergence}. No such results exist for other BNSP models. It would be interesting to see how posterior concentration analysis could be integrated into the hierarchical partition or even bounding-based partition BNSP models.

\subsection{Analysis of deep neural networks}
Recent work~\cite{balestriero2018spline,NEURIPS2019_0801b20e} explores the possibility of using a Power-Diagram partition -- a generalisation of a Voronoi tiling partition -- to analyse the piecewise-affine activation functions between the layers  of a Deep Neural Network (DNN). However, this work is restricted to analysing piecewise-affine activation functions and is unable to be applied to general DNN architecture with non-linear activation functions. 
Exploring these ideas with more flexible and advanced partitioning structures could bring great insights and new ideas to DNN research.

\section{Conclusion}
In this survey, we provide a first review on the Bayesian Nonparametric Space Partitioning~(BNSP) models, through studying five typical BNSP strategies, detailing the current applications of BNSP models and discussing the current challenges of BNSP models. 

{\small \bibliographystyle{named}
\bibliography{Xuhui_Machine_Learning}}

\begin{thebibliography}{}

\bibitem[\protect\citeauthoryear{Aldous}{1981}]{aldous1981representations}
David~J. Aldous.
\newblock Representations for partially exchangeable arrays of random
  variables.
\newblock {\em Journal of Multivariate Analysis}, 11(4):581--598, 1981.

\bibitem[\protect\citeauthoryear{Balestriero and
  Baraniuk}{2018}]{balestriero2018spline}
Randall Balestriero and Richard Baraniuk.
\newblock A spline theory of deep learning.
\newblock In {\em ICML}, pages 374--383, 2018.

\bibitem[\protect\citeauthoryear{Balestriero \bgroup \em et al.\egroup
  }{2019}]{NEURIPS2019_0801b20e}
Randall Balestriero, Romain Cosentino, Behnaam Aazhang, and Richard Baraniuk.
\newblock The geometry of deep networks: Power diagram subdivision.
\newblock In {\em NeurIPS}, pages 15832--15841, 2019.

\bibitem[\protect\citeauthoryear{Balog \bgroup \em et al.\egroup
  }{2016}]{balog2016mondrian}
Matej Balog, Balaji Lakshminarayanan, Zoubin Ghahramani, Daniel~M Roy, and
  Yee~Whye Teh.
\newblock The mondrian kernel.
\newblock {\em arXiv preprint arXiv:1606.05241}, 2016.

\bibitem[\protect\citeauthoryear{Breiman}{2000}]{breiman2000some}
Leo Breiman.
\newblock Some infinity theory for predictor ensembles.
\newblock Technical report, Technical Report 579, Statistics Dept. UCB, 2000.

\bibitem[\protect\citeauthoryear{Caldas and Kaski}{2008}]{caldas2008bayesian}
Jos{\'e} Caldas and Samuel Kaski.
\newblock Bayesian biclustering with the plaid model.
\newblock In {\em MLSP 2008. IEEE Workshop on}, pages 291--296. IEEE, 2008.

\bibitem[\protect\citeauthoryear{Chipman \bgroup \em et al.\egroup
  }{2010}]{chipman2010bart}
Hugh~A. Chipman, Edward~I. George, and Robert~E. McCulloch.
\newblock Bart: Bayesian additive regression trees.
\newblock {\em The Annals of Applied Statistics}, 4(1):266--298, 2010.

\bibitem[\protect\citeauthoryear{Dulucq and Guibert}{1998}]{DULUCQ1998143}
S.~Dulucq and O.~Guibert.
\newblock Baxter permutations.
\newblock {\em Discrete Mathematics}, 180(1):143--156, 1998.

\bibitem[\protect\citeauthoryear{Fan \bgroup \em et al.\egroup
  }{2018a}]{pmlr-v84-fan18b}
Xuhui Fan, Bin Li, and Scott~A. Sisson.
\newblock The binary space partitioning-tree process.
\newblock In {\em AISTATS}, volume~84, pages 1859--1867, 2018.

\bibitem[\protect\citeauthoryear{Fan \bgroup \em et al.\egroup
  }{2018b}]{NIPS2018_RBP}
Xuhui Fan, Bin Li, and Scott~A. Sisson.
\newblock Rectangular bounding process.
\newblock In {\em NeurIPS}, pages 7631--7641, 2018.

\bibitem[\protect\citeauthoryear{Fan \bgroup \em et al.\egroup
  }{2019}]{pmlr-v89-fan18a}
Xuhui Fan, Bin Li, and Scott~A. Sisson.
\newblock Binary space partitioning forests.
\newblock In {\em AISTATS}, volume~89, pages 3022--3031, 2019.

\bibitem[\protect\citeauthoryear{Fan \bgroup \em et al.\egroup
  }{2020}]{fan2020onlinebspf}
Xuhui Fan, Bin Li, and Scott~A. Sisson.
\newblock Online binary space partitioning forests.
\newblock In {\em AISTATS}, 2020.

\bibitem[\protect\citeauthoryear{Ferguson}{1973}]{ferguson1973bayesian}
Thomas~S Ferguson.
\newblock A bayesian analysis of some nonparametric problems.
\newblock {\em The Annals of Statistics}, pages 209--230, 1973.

\bibitem[\protect\citeauthoryear{Ge \bgroup \em et al.\egroup
  }{2019}]{random_tessellation_forests}
Shufei Ge, Shijia Wang, Yee~Whye Teh, Liangliang Wang, and Lloyd Elliott.
\newblock Random tessellation forests.
\newblock In {\em NeurIPS}, pages 9571--9581, 2019.

\bibitem[\protect\citeauthoryear{Ghosal \bgroup \em et al.\egroup
  }{2007}]{ghosal2007convergence}
Subhashis Ghosal, Aad Van Der~Vaart, et~al.
\newblock Convergence rates of posterior distributions for noniid observations.
\newblock {\em The Annals of Statistics}, 35(1):192--223, 2007.

\bibitem[\protect\citeauthoryear{Givoni \bgroup \em et al.\egroup
  }{2006}]{Givoni06}
Inmar Givoni, Vincent Cheung, and Brendan Frey.
\newblock Matrix tile analysis.
\newblock In {\em UAI}, pages 200--207, 2006.

\bibitem[\protect\citeauthoryear{Ishiguro \bgroup \em et al.\egroup
  }{2010}]{ishiguro2010dynamic}
Katsuhiko Ishiguro, Tomoharu Iwata, Naonori Ueda, and Joshua~B. Tenenbaum.
\newblock Dynamic infinite relational model for time-varying relational data
  analysis.
\newblock In {\em NIPS}, pages 919--927, 2010.

\bibitem[\protect\citeauthoryear{Kemp \bgroup \em et al.\egroup
  }{2006}]{kemp2006learning}
Charles Kemp, Joshua~B. Tenenbaum, Thomas~L. Griffiths, Takeshi Yamada, and
  Naonori Ueda.
\newblock Learning systems of concepts with an infinite relational model.
\newblock In {\em AAAI}, volume~3, pages 381--388, 2006.

\bibitem[\protect\citeauthoryear{Kingma and Welling}{2013}]{kingma2013auto}
Diederik~P Kingma and Max Welling.
\newblock Auto-encoding variational bayes.
\newblock {\em arXiv preprint arXiv:1312.6114}, 2013.

\bibitem[\protect\citeauthoryear{Kuck \bgroup \em et al.\egroup
  }{2019}]{NIPS2019_9089}
Jonathan Kuck, Tri Dao, Hamid Rezatofighi, Ashish Sabharwal, and Stefano Ermon.
\newblock Approximating the permanent by sampling from adaptive partitions.
\newblock In {\em NeurIPS}, pages 8860--8871. 2019.

\bibitem[\protect\citeauthoryear{Lakshminarayanan \bgroup \em et al.\egroup
  }{2014}]{LakRoyTeh2014a}
Balaji Lakshminarayanan, Daniel~M. Roy, and Yee~Whye Teh.
\newblock {Mondrian} forests: Efficient online random forests.
\newblock In {\em NIPS}, pages 3140--3148, 2014.

\bibitem[\protect\citeauthoryear{Lakshminarayanan \bgroup \em et al.\egroup
  }{2016}]{lakshminarayanan2016mondrian}
Balaji Lakshminarayanan, Daniel~M Roy, and Yee~Whye Teh.
\newblock Mondrian forests for large-scale regression when uncertainty matters.
\newblock In {\em AISTATS}, pages 1478--1487, 2016.

\bibitem[\protect\citeauthoryear{Mourtada \bgroup \em et al.\egroup
  }{2017}]{consistencyMondrianforest}
Jaouad Mourtada, St\'{e}phane Ga\"{\i}ffas, and Erwan Scornet.
\newblock Universal consistency and minimax rates for online mondrian forests.
\newblock In {\em NIPS}, pages 3758--3767, 2017.

\bibitem[\protect\citeauthoryear{Mourtada \bgroup \em et al.\egroup
  }{2018}]{mourtada2018minimax}
Jaouad Mourtada, St{\'e}phane Ga{\"\i}ffas, and Erwan Scornet.
\newblock Minimax optimal rates for mondrian trees and forests.
\newblock {\em arXiv preprint arXiv:1803.05784}, 2018.

\bibitem[\protect\citeauthoryear{Nakano \bgroup \em et al.\egroup
  }{2014a}]{nakano2014rectangular}
Masahiro Nakano, Katsuhiko Ishiguro, Akisato Kimura, Takeshi Yamada, and
  Naonori Ueda.
\newblock Rectangular tiling process.
\newblock In {\em ICML}, pages 361--369, 2014.

\bibitem[\protect\citeauthoryear{Nakano \bgroup \em et al.\egroup
  }{2014b}]{mhmm2014nakano}
Masahiro Nakano, Yasunori Ohishi, Hirokazu Kameoka, Ryo Mukai, and Kunio
  Kashino.
\newblock Mondrian hidden {Markov} model for music signal processing.
\newblock In {\em ICASSP}, pages 2405--2409, 2014.

\bibitem[\protect\citeauthoryear{Nakano \bgroup \em et al.\egroup
  }{2020}]{NEURIPS2020_6271faad}
Masahiro Nakano, Akisato Kimura, Takeshi Yamada, and Naonori Ueda.
\newblock Baxter permutation process.
\newblock In {\em NeurIPS}, pages 8648--8659, 2020.

\bibitem[\protect\citeauthoryear{Nowicki and
  Snijders}{2001}]{nowicki2001estimation}
Krzysztof Nowicki and Tom~A.B. Snijders.
\newblock Estimation and prediction for stochastic block structures.
\newblock {\em Journal of the American Statistical Association},
  96(455):1077--1087, 2001.

\bibitem[\protect\citeauthoryear{Oksendal}{2013}]{oksendal2013stochastic}
Bernt Oksendal.
\newblock {\em Stochastic differential equations: an introduction with
  applications}.
\newblock Springer Science \& Business Media, 2013.

\bibitem[\protect\citeauthoryear{Orbanz and Roy}{2015}]{TPAMI2014peterdaniel}
Peter Orbanz and Daniel~M. Roy.
\newblock {Bayesian} models of graphs, arrays and other exchangeable random
  structures.
\newblock {\em IEEE Transactions on Pattern Analysis and Machine Intelligence},
  37(02):437--461, 2015.

\bibitem[\protect\citeauthoryear{O'Reilly and Tran}{2020}]{o2020stochastic}
Eliza O'Reilly and Ngoc Tran.
\newblock Stochastic geometry to generalize the mondrian process.
\newblock {\em arXiv preprint arXiv:2002.00797}, 2020.

\bibitem[\protect\citeauthoryear{Rahimi and Recht}{2008}]{rahimi2008random}
Ali Rahimi and Benjamin Recht.
\newblock Random features for large-scale kernel machines.
\newblock In {\em NIPS}, pages 1177--1184, 2008.

\bibitem[\protect\citeauthoryear{Rockov{\'a} and van~der
  Pas}{2017}]{rockova2017posterior}
Veronika Rockov{\'a} and St{\'e}phanie van~der Pas.
\newblock Posterior concentration for bayesian regression trees and forests.
\newblock {\em arXiv preprint arXiv:1708.08734}, 2017.

\bibitem[\protect\citeauthoryear{Roy and Teh}{2009}]{roy2009mondrian}
Daniel~M. Roy and Yee~Whye Teh.
\newblock The {Mondrian} process.
\newblock In {\em NIPS}, pages 1377--1384, 2009.

\bibitem[\protect\citeauthoryear{Schmidt and M{\o}rup}{2013}]{nonpa2013schmidt}
Mikkel~N. Schmidt and Morten M{\o}rup.
\newblock Nonparametric {Bayesian} modeling of complex networks: An
  introduction.
\newblock {\em IEEE Signal Processing Magazine}, 30(3):110--128, 2013.

\bibitem[\protect\citeauthoryear{Sethuraman}{1994}]{sethuraman1994constructive}
Jayaram Sethuraman.
\newblock A constructive definition of dirichlet priors.
\newblock {\em Statistica sinica}, pages 639--650, 1994.

\bibitem[\protect\citeauthoryear{Wang \bgroup \em et al.\egroup
  }{2011}]{wang2011nonparametric}
Pu~Wang, Kathryn~B. Laskey, Carlotta Domeniconi, and Michael~I. Jordan.
\newblock Nonparametric {Bayesian} co-clustering ensembles.
\newblock In {\em SDM}, pages 331--342, 2011.

\end{thebibliography}

\end{document}